\documentclass{article}

\usepackage{microtype}
\usepackage{graphicx}
\usepackage{subfigure}
\usepackage{booktabs} 

\usepackage{hyperref}

\usepackage[accepted]{icml2025}

\usepackage{amsmath}
\usepackage{amssymb}
\usepackage{mathtools}
\usepackage{amsthm}
\usepackage{graphicx}
\usepackage{amsmath}
\usepackage{amsthm}
\usepackage{booktabs}
\usepackage{algorithm}
\usepackage{algorithmic}
\usepackage[switch]{lineno}
\usepackage{amsthm,amsmath,amssymb}
\renewcommand{\algorithmicrequire}{\textbf{Input:}}
\renewcommand{\algorithmicensure}{\textbf{Output:}}
\usepackage{mathrsfs}
\renewcommand{\algorithmiccomment}[1]{\hfill $\triangleright$ #1}
\usepackage{colortbl}
\usepackage{color}
\usepackage{multirow}
\usepackage[capitalize,noabbrev]{cleveref}

\theoremstyle{plain}

\theoremstyle{definition}

\theoremstyle{remark}

\newcommand{\RETURN}{\textbf{return }}
\usepackage[textsize=tiny]{todonotes}

\begin{document}

\twocolumn[
\icmltitle{A General Error-Theoretical Analysis Framework for Constructing Compression Strategies}



\icmlsetsymbol{equal}{*}

\begin{icmlauthorlist}
\icmlauthor{Boyang Zhang}{comp,yyy,sch}
\icmlauthor{Daning Cheng*}{yyy}
\icmlauthor{Yunquan Zhang}{yyy}
\icmlauthor{Meiqi Tu}{schku}
\icmlauthor{Fangming Liu}{sch}
\icmlauthor{Jiake Tian}{schu}
zhangby01@pcl.ac.cn, \{chengdaning, zyq\}@ict.ac.cn, \\  tumeiqi24@connect.hku.hk, fangminghk@gmail.com, mijiake@mail.scut.edu.cn
\end{icmlauthorlist}

\icmlaffiliation{yyy}{Institute of Computing Technology, Chinese Academy of Sciences, Beijing, China}
\icmlaffiliation{comp}{University of Chinese Academy of Sciences, Beijing, China}
\icmlaffiliation{sch}{Peng Cheng Laboratory, Shenzhen, China}
\icmlaffiliation{schu}{the School of Microelectronics, South China University of Technology, Guangzhou, China}
\icmlaffiliation{schku}{The University of Hong Kong}
\icmlcorrespondingauthor{}

\icmlkeywords{Machine Learning, ICML}

\vskip 0.3in
]



\printAffiliationsAndNotice{}  

\begin{abstract}
The exponential growth in parameter size and computational complexity of deep models poses significant challenges for efficient deployment. The core problem of existing compression methods is that different layers of the model have significant differences in their tolerance to compression levels. For instance, the first layer of a model can typically sustain a higher compression level compared to the last layer without compromising performance. Thus, the key challenge lies in how to allocate compression levels across layers in a way that minimizes performance loss while maximizing parameter reduction.
To address this challenge, we propose a Compression Error Theory (CET) framework, designed to determine the optimal compression level for each layer. Taking quantization as an example, CET leverages differential expansion and algebraic geometry to reconstruct the quadratic form of quantization error as ellipsoids and hyperbolic paraboloids, and utilizes their geometric structures to define an error subspace. To identify the error subspace with minimal performance loss, by performing orthogonal decomposition of the geometric space, CET transforms the optimization process of the error subspace into a complementary problem. The final theoretical analysis shows that constructing the quantization subspace along the major axis results in minimal performance degradation. 
Through experimental verification of the theory, CET can greatly retain performance while compressing. Specifically, on the ResNet-34 model, CET achieves nearly 11$\times$ parameter compression while even surpassing performance comparable to the original model. 
\end{abstract}

\vspace{-0.5cm}
\section{Introduction}
Existing research shows that increasing model size and training data can significantly enhance the performance and learning capability of deep models. However, such improvements often come with a sharp increase in computational complexity and storage requirements, especially in resource-constrained hardware and latency-sensitive scenarios. Thus, a critical challenge is how to efficiently compress model parameters while maintaining nearly unchanged performance. 

Mixed compression techniques have become a mainstream approach for model compression. Due to the varying contributions of different layers to overall model performance, compared with the unified compression approach, different layers exhibit significantly different tolerance levels to compression. Adopting layer-wise differentiated compression strategies can thus maximize compression rates while minimizing performance degradation. For example, in low-rank decomposition, different layers have varying ranks. In quantization, some critical layers require higher precision, while others can use lower precision. However, the main challenge with this approach lies in the exponentially growing search space for determining the optimal mixed-precision quantization configuration. Specifically, for a neural network with $L$ layers, where each layer can select from four possible bit-widths (e.g., 2/3/4/8 bits), the search space grows to $4^L$. The huge search space makes it almost impossible to find the optimal configuration that can maintain good generalization performance and meet hardware efficiency.

Several representative compression methods \cite{zhao2021distribution, frantar2022gptq, dong2019hawq} have been proposed to address these challenges. For instance, DMBQ pre-emptively searches for the optimal bit-width configuration in the distribution space and dynamically selects compression settings during training. The HAWQ series computes layer-wise Hessian information to assess the relative sensitivity of each layer, thereby determining the optimal compression configuration. However, these approaches suffer from the following limitations: (1) Lack of optimality explanation: Existing compression configurations often rely on heuristic approaches, lacking a clear theoretical foundation to justify their optimality. (2) Neglect of error correlation: These methods typically decouple the optimization of model performance error and compression error, failing to systematically analyze their theoretical interdependence. (3) Limited adaptability to aggressive quantization: At lower bit widths, these methods struggle to handle degradation caused by compression errors, which can disrupt retraining and result in significant performance drops.

\begin{figure*}
\centering
\includegraphics[width=16cm,height=5.0cm]{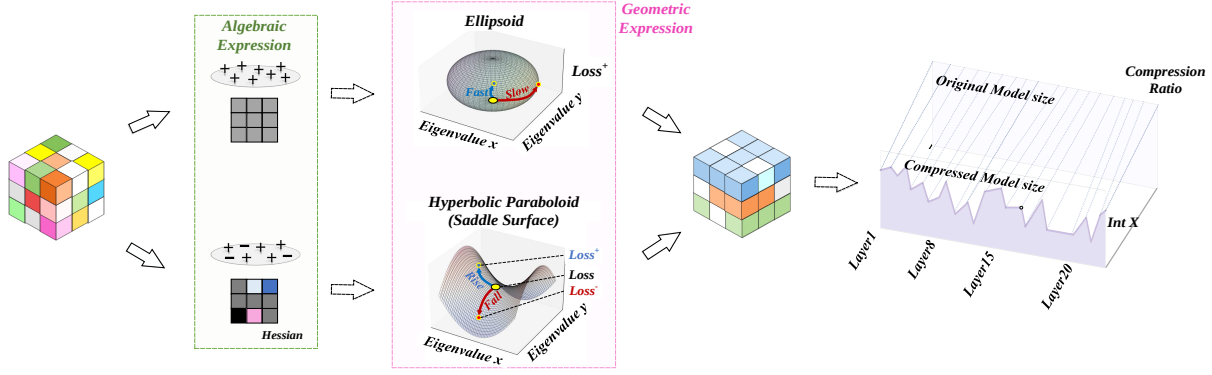}
\vspace{-0.4cm}
\caption{The CET framework completes the transition from algebraic to geometric analysis, providing a theoretical foundation for the quantization error vector. Based on the positive definiteness of the Hessian matrix at the convergence point, CET reconstructs the quadratic form of quantization error into ellipsoids or hyperbolic paraboloids. For ellipsoids, the direction along the long axis corresponds to the direction with the slowest increase in loss. For hyperbolic paraboloids, analogous to ellipsoids, the eigenvector corresponding to the negative eigenvalue defines the long-axis direction, representing the direction of loss reduction. Theoretical analysis suggests that the quantization error vector should be optimized along these two directions.}
\vspace{-0.4cm}
\label{fig1}
\end{figure*}

To address the aforementioned challenges, we propose a general Compression Error Theory (CET) framework, which systematically derives the optimal compression configuration. As shown in Figure \ref{fig1}, taking quantization as an example, CET first establishes the relationship between quantization-induced parameter errors and performance loss errors using total differentiation, and then applies algebraic geometry to transform the quadratic form of loss errors into a geometric representation. This geometric representation directly guides the selection of the quantization parameter space, clearly identifying the optimal quantization direction and range, and is more adaptable to low-bit-width compression. Next, by utilizing the theory of orthogonal complements, the process of solving the error subspace is converted into a complementary problem, allowing for the determination of the best quantization configuration. Unlike traditional methods, CET does not require retraining and can directly achieve the optimal compression configuration. Experimental results show that CET can successfully maximize model compression with minimal performance degradation. Specifically, for the ResNet-34 model, CET achieves nearly 11 parameter compression while even surpassing performance comparable to the original model. Existing compression error analysis methods rarely explore the problem from the perspective of algebraic geometry and spatial structure. Our approach leverages algebraic geometry and total differential analysis to examine the geometric properties and spatial structure of parameters during the compression process, using these insights to guide the direction and magnitude of compression. Although CET is based on quantization, its applicability extends beyond quantization and can be widely applied to other compression methods. 
The main contributions are as follows:

\textbf{$1)$} We propose a general Compression Error Theory (CET) framework, which derives the optimal compression configuration through theoretical analysis. CET is independent of specific compression methods and has broad applicability.

\textbf{$2)$} The proposed CET combines total differentiation and algebraic geometry to precisely guide the selection of the compression parameter space, avoiding the need for retraining commonly required by traditional methods.

\textbf{$3)$} Experimental results show that CET can maximize model compression with minimal performance degradation. For instance, on the ResNet-34 model, CET achieves nearly 11 parameter compression while even surpassing performance comparable to the original model.

\vspace{-0.4cm}
\section{Related Works}
\subsection{Neural Network in Function}
We present the analysis of neural networks as composite functions. All our conclusions are independent of the structure of the neural network.
First, for an n-layer neural network model, the loss of the model is optimized according to the following equation
\begin{small} 
\begin{equation}
\begin{aligned} & \begin{aligned}
\min_{\textbf{W}}f(\textbf{W})=\mathbf{E}_{Sample}\ell(\textbf{W},Sample)=\frac{1}{m}\sum_{(x_{i},y_{i})\in\mathbb{D}}\ell(\textbf{W},x_{i},y_{i})\end{aligned},\\
&\ell(\textbf{W},x_{i},y_{i})=L(model_{n}(x_{i},\textbf{W}),y_{i}),\\  & model_{n}=h_1(h_2(h_3(h_4(\cdots(h_{n+1},w_{n})\cdots,w_4),w_3),w_2),w_1),\end{aligned}
\label{eq1}
\end{equation}\end{small}

\noindent where $f(\cdot)$ represents the loss of the model on a dataset, $\mathbf{E}$ stands for expectation, $m$ is the size of the dataset, $\ell(\cdot)$ is the loss function for a sample, and $(x_i,y_i)$ denotes a sample in the dataset along with its corresponding label, $L(\cdot)$ represents the loss function, such as the cross-entropy function; $h_i$, with $i\in[1,...,n]$ represents the ($n-i+1$)th layer in the neural network, $\textbf{W} = (w_n^T,w_{n-1}^T,\cdots,w_1^T)^T$, where $w_i$ is the parameter in $h_i(\cdot)$, and for the reason of a unified format, $h_{n+1}$ denotes the sample $x$.
When the model is treated as a complex high-dimensional nonlinear mapping, it encapsulates the structural constraints of the network and the characteristics of the loss function. Its local properties can be studied through differential and algebraic geometry methods to uncover the local shape characteristics of the function.
\subsection{Quantization for Compression}
The computational units of deep models are primarily composed of matrix multiplication. Quantization accelerates the multiplication process by converting floating-point parameters into lower-bit formats, thus speeding up the inference process. For a single layer of a neural network, it is represented as $Y=X\cdot W \in \mathbb{R}^{S\times C_{out}}$, where $X\in \mathbb{R}^{S\times C_{in}}$ is the activation input and $W \in \mathbb{R}^{C_{in}\times C_{out}}$ is the weight matrix. Taking integer uniform quantization as an example, the $b$-bit quantization process maps the FP16/32 weight tensor $W$ to a lower-bit integer $W_q$.
\begin{equation}
\begin{aligned}
Q(\textbf{W})&=\mathrm{clamp}\left(\left\lfloor\frac{\mathbf{W}}{\Delta}\right\rceil+z,0,2^b-1\right) \\
\mathrm{where}\quad\Delta&=\frac{\max(\mathbf{W})-\min(\mathbf{W})}{2^b-1},z=-\left\lfloor\frac{\min(\mathbf{W})}{\Delta}\right\rceil
\end{aligned}
\label{quant}
\end{equation}
The notation is $\lfloor\cdot\rceil$ means the nearest rounding operation, $\Delta$ is the quantization step size and $z$ represents the zero point.
When adopting minimum square error (MSE) as the criterion, the quantization process is expressed as the following minimization error problem:
\begin{equation}
\operatorname*{min}\|\textbf{W}-Q(\textbf{W})\|_{2}\quad s.t.Q(\textbf{W})\in\Pi_{b}
\end{equation}
$Q(\textbf{W})$:$\mathbb{R}^D\times\mathbb{Z}^+\to\Pi_b$ is the quantization function (Equation \ref{quant}).
Existing methods focus on reducing parameter errors through hybrid quantization schemes, which can be divided into two categories: search-based and tolerance-based methods.
Search-based methods, such as those by \cite{wang2019haq, lou2019autoq, wu2018mixed} treat the quantized network as a whole and use various bit-width allocations to evaluate the model. The evaluation results are then used to guide the search process to find the optimal solution. However, these methods are computationally expensive, difficult to parallelize, and due to their iterative search nature, require hundreds or thousands of GPU hours.

To optimize efficiency, another category of tolerance-based methods measures each layer's tolerance to quantization errors. When the tolerance of a layer is higher, the layer can be quantized with a lower bit-width. Various sensitivity metrics have been proposed in practice, such as the Kullback-Leibler divergence between the quantized layer output and the full-precision layer output (\cite{cai2020zeroq}), the maximum eigenvalue of the Hessian (\cite{2019HAWQ}), the trace of the Hessian (\cite{dong2019hawq, 2020HAWQV3}), the Gaussian-Newton matrix approximation of the Hessian (\cite{chen2021towards}), or the quantization factor (\cite{tang2022mixed}). All tolerance-based methods minimize the sum of tolerance across layers under the constraint of target compression ratio.
Although these methods are effective in practice, they lack a theoretical foundation to justify the optimality of their results. Furthermore, since these methods do not consider the relationship between model performance and parameter errors, they struggle to address model degradation caused by compression errors at lower bit widths, which lead to retraining failures or sharp performance drops.
\vspace{-0.4cm}
\section{Compression Error Theoretical Analysis Framework}
\subsection{Error Correlation and Optimization}
Typically, compression involves two types of errors: the change in parameters after compression, $\Delta_w$, and the change in loss caused by the parameter variation, $\Delta_L$. CET first establishes the relationship between the two and performs a unified analysis, aiming to determine the direction and magnitude of $\Delta_w$ to minimize $\Delta_L$. 

The mathematical essence of many compression schemes such as quantization and decomposition is to introduce compression errors into the original parameters. After compression, for a sample, the model loss $\bar{\ell}$ during inference is reformulated as the following equation
\begin{small}
\begin{align}
\bar{\ell}_k(w,x_{j},y_{j})  =L(h_1(h_2(\cdots h_{n}(x_i, w_{n}+\delta^k_{n})\cdots,\nonumber  \\w_2+\delta^k_2),w_1+\delta^k_1), y_{i}),
\end{align}
\end{small}  
where $\delta^k_{i}\in\Delta_w,i\in\left\{1,\cdots,n\right\}$ denotes the noise error on the weights after the k-level compression. In quantization, the k-level represents different bit widths.
CET directly associates the compression error and the change of the loss function through total differentials.
According to total differentials, the following equation can be obtained
\begin{equation}
	\begin{aligned}
		\Delta_L = \bar{\ell}(w,x_i,y_i)-\ell(w,x_i,y_i) =  \sum_{i=1}^n\frac{\partial\ell}{\partial w_i}\cdot\delta_i+ & \\ \frac{1}{2} \delta_i^T\mathbb{H}\delta_i+O(||\delta_i||^n)
	\end{aligned} \label{eq5}
 \end{equation}
where $\mathbb{H}$ represents the Hessian matrix and $O(||(\delta_i)||^n) $ represents the high-order term, $\cdot$ is inner product. For the loss on the whole dataset, we can gain
\begin{equation}
	\begin{aligned}
		\min\limits_{\delta\in \Delta}\Delta_L=\min\limits_{\delta\in \Delta}\bar{f}(w)-f(w)=\frac{1}{m}\sum\limits_{(x_j,y_j)\in\mathbb{D}}\sum\limits_{i=1}^{n}\frac{\partial\ell}{\partial w_i}\cdot\delta_i \\ +\frac{1}{2}\delta_i^T\mathbb{H} \delta_i+O(||\delta_i||^n)
	\end{aligned} \label{eq6}
 \end{equation}
where $\bar{f}(w) = \frac{1}{m}\sum \bar{\ell}(\cdot)$. This equation directly links compression $\Delta_w$ and model performance $\Delta_L$.
Although higher-order differentials provide theoretical support for CET, their usage requires meeting the following conditions. First, the function must be smooth and differentiable; second, the parameter changes must be small enough. According to the chain rule, multi-layer neural networks are continuously differentiable concerning all parameters, meaning that they are inherently smooth and differentiable. Thus, Eq. \ref{eq5} generally satisfies $C^k$ continuity. Since the scale of compression determines the parameter variation, we primarily focus on the magnitude of the error. When the variable $\delta$ is sufficiently small, the actual change in the loss function can be accurately described by the total differential 
$df$. Therefore, determining the "sufficiently small" threshold in the practical model is crucial. Since each layer can accommodate different sizes of parameter errors, we compute the gap between theory and practice, denoted as $U(x)$.
\begin{small} 
\begin{equation}
\begin{aligned}
U_{\delta^k}\left(x_i\right): |{\ell}(w \pm \delta^k_{i} ,x_{i},y_{i}) -(\ell(w,x_{i},y_{i})+\sum_{i=1}^{n}\frac{\partial\ell_k}{\partial w_i}\cdot\delta^k_{i})+& \\ \frac{1}{2}\delta_i^T\mathbb{H}\delta_i+O(||(\delta_i)||^n)|
\end{aligned}\label{eq7}
\end{equation}
\end{small}

The left-hand side of the equation represents the loss caused by actual noise interference, while the right-hand side represents the theoretical loss caused by noise. The parameter $k$ controls the compression level. When the neighborhood of compression error is smaller than $10^{-3}$, we consider the actual error to be close to the theoretical error. For weights, ideally, the first-order term in a well-trained model should be zero. Since higher-order terms are uncomputable, the impact of the second-order term is typically considered. Hence, the update for the optimization term is given by the following equation,

\begin{equation}
\begin{aligned}
\min\limits_{\delta\in \Delta}\Delta_L = \bar{f}(w)-f(w)=\frac{1}{2}\delta_i^T\mathbb{H} \delta_i
\end{aligned} \label{eq8}
\end{equation}
$\delta_i^T\mathbb{H} \delta_i$ is a quadratic expression, and $\mathbb{H}$ is composed of the second-order derivatives of the whole model. We hope that the loss decreases or increases small and slowly after compression. Next, CET mainly uses algebraic geometry to analyze this quadratic expression.

\subsection{Reconstruction of the Compression Subspace}
The expression $\delta_i^T\mathbb{H} \delta_i$ serves as an abstract representation of the quadratic term of a function, which can describe geometric surfaces in high-dimensional space, where $\mathbb{H}$ acts as the coefficient matrix. Its fundamental geometric form is given by: 
\begin{equation}
\delta_i^T\mathbb{H} \delta_i = c \label{eq9}
\end{equation}
where $c$ is a constant that represents the isosurface of the geometric shape. This equation constrains the weight vector $w$ within an n-dimensional space, where $n$ is the dimensionality of the weight vector. As shown in Figure \ref{fig1},
the eigenvalues and eigenvectors of $\mathbb{H}$ define the geometric properties of the surface: the eigenvalues indicate the degree of stretching or compression along the principal axes, while the eigenvectors determine the orientation of these axes. Furthermore, the definiteness of $\mathbb{H}$ dictates the global shape of the surface. Specifically, a positive definite matrix corresponds to a closed surface (e.g., an ellipsoid), whereas an indefinite matrix may result in an open surface (e.g., a hyperbolic paraboloid) \cite{hartshorne2013algebraic}. Hence, a key step in CET is to determine the overall geometric shape of the surface, which serves as the foundation for guiding the direction and magnitude of model compression.
We use the eigendecomposition to perform a standard form transformation on Eq.\ref{eq9}.
\begin{equation}
Q(\mathbf{y})=\lambda_1y_1^2+\lambda_2y_2^2+\cdots+\lambda_ny_n^2 \label{eq10}
\end{equation}
where $y=P^Tw$ is the new coordinate system after the eigenvector transformation. $P$ is an orthogonal matrix composed of $\mathbb{H}$(eigenvectors). The quadratic expression is a weighted sum in the direction of each principal axis, and its geometric shape is completely determined by the eigenvalues and eigenvectors. When $\mathbb{H}$ is positive definite at the convergence point $\textbf{W}$, the overall shape of the level surface 
$c$ forms a closed ellipsoid, indicating that the convergence point is locally convex. Conversely, if $\mathbb{H}$ is indefinite at the convergence point, the level surface becomes an open hyperbolic paraboloid. Negative definiteness at the convergence point, where the level surface would exhibit a fully concave open structure corresponding to a local maximum, is not possible in this context. This ensures that such cases are excluded.

Given that our objective is to minimize or slow the increase in loss after compression, the compression vector should align as closely as possible with the direction of the long axis of the ellipsoid or hyperbolic paraboloid. In this direction, the curvature is smaller, the changes in the level surface are more gradual, and the compression vector has a minimal impact on the loss value. Figure \ref{fig1} illustrates the transition of CET from a quadratic algebraic representation to a geometric interpretation, defining the optimization path for the compression vector through geometry. Next, after determining the direction of the quantized subspace, the next step is how to efficiently solve this space.

\subsection{Solution of the Compression Subspace}
To construct the subspace of the long axis, we leverage the concept of complementary spaces. In the $\mathbb{R}^n$ space defined by the curvature of the surface, the Hessian matrix's eigenvectors form a complete orthogonal basis. This allows us to decompose $\mathbb{R}^n$ into two complementary subspaces, satisfying the following equation
\begin{equation}
\mathbb{R}^n=V_{\mathrm{long}}\oplus V_{\mathrm{short}} \label{eq11}
\end{equation}
Here, $\oplus$ denotes the direct sum relationship, where $V_{\mathrm{long}}$ represents the subspace of the long axis, and $V_{\mathrm{short}}$ represents the subspace of the short axis.
Our goal is to find a compression vector that resides in the long-axis subspace. To achieve this, CET reformulates the problem by solving for the zero space of the short-axis subspace. This approach effectively identifies vectors orthogonal to the short-axis subspace and hence aligns with the long-axis subspace.
\begin{equation}
\begin{cases}
 \lambda_1 y^2_1(\delta_i) = 0, \\
\lambda_2 y^2_2(\delta_i) = 0, \\
\vdots \\
\lambda_m y^2_m(\delta_i) = 0
\end{cases} \label{eq12}
\quad i\in\{1, 2, ...,n\}
\end{equation}
where $\lambda_m$ represents the eigenvalues corresponding to the short-axis subspace, and $y_m$ denotes the transformed eigenvectors associated with these eigenvalues and noise $\delta_i$ on the weights. $n$ (the number of parameters) is much larger than $m$ (the number of eigenvalues), which means that this is an indeterminate system of equations. By constructing the zero space of the short-axis subspace, CET isolates vectors orthogonal to $V_{\mathrm{short}}$, thereby enabling the identification of compression vectors in the long-axis subspace.

The solution to the indeterminate equation system is ill-posed, thus requiring further constraints on Eq.\ref{eq12}:
\begin{itemize}
\item The ideal solution to Eq.\ref{eq12} is that every term $y$ equals zero, which corresponds to no quantization. This result is intuitive, as the absence of quantization minimizes parameter error. However, CET avoids this trivial solution by constraining the model size. Specifically, the condition Modelsize $M_{compress} < M_{orgin}$ is introduced, ensuring that the parameter error is strictly non-zero and the model volume is effectively compressed.
\item When the solved parameter error $\delta$ becomes excessively large, Eq.\ref{eq8} indicates that the loss change $\Delta_L$ will also increase significantly. Thus, we not only require the compression vector to exist within the long-axis subspace but also minimize its magnitude. A smaller magnitude implies reduced compression loss along the long-axis direction. CET imposes an additional minimization constraint on the parameter error, namely $min(\Vert \delta_i \Vert ^2)$. 
\end{itemize}
With this refinement, Eq.\ref{eq12} is updated to the following form:
\begin{equation}
\begin{cases} 
\lambda_1 y_1^2 (\delta_i) = 0, \\
\lambda_2 y_2^2 (\delta_i) = 0, \\
\vdots \\
\lambda_m y_m^2 (\delta_i) = 0
\end{cases}
\quad \text{s.t.} \quad 
\begin{cases} 
M_\text{compress} < M_{\text{origin}}, \\
\|\delta_i\|^2 < \epsilon. 
\end{cases}
 \label{eq13}
\end{equation}

where $\epsilon$ approaches 0, $i\in\{1, 2, ...,n\}$. By solving the above system of equations, we can directly obtain the parameter error $\Delta_w$ that minimizes $\Delta_L$, thereby determining the level of model compression.

\begin{algorithm}[!ht]
\caption{CET Algorithm for Quantization}
\begin{algorithmic}[1]
\REQUIRE Convergence point parameters $w$, calibration dataset $\mathcal{D}$, loss function $L(w)$.
\ENSURE Quantization bit-widths for each layer.
\STATE \textbf{Step 1: Compute Hessian matrix eigenvalues $\lambda$ and eigenvectors $v_i$ using the Lanczos algorithm on the calibration dataset $\mathcal{D}$}
\begin{itemize}
    \item Start with a random initial vector $v_1$ and iteratively perform matrix-vector multiplications with $H= \frac{\partial^2 L(w)}{\partial w^2}$.
    \item Orthogonalize the vectors to build a Krylov subspace and construct a tridiagonal matrix $T_k$.
    \item Solve $T_k$ to obtain approximate eigenvalues $\lambda_1, \lambda_2, \dots, \lambda_k$ and eigenvectors.
\end{itemize}
\STATE Return $\lambda$ and the corresponding eigenvectors $v_i$.

\STATE \textbf{Step 2: Randomly initialize perturbation $\delta$}
\STATE Randomly initialize $\delta_0\sim \mathcal{N}(0, \sigma^2)$ as the starting point for optimization. \COMMENT {Integration of multiple sampling results.}

\STATE \textbf{Step 3: Construct $y$ via canonical transformation}
\STATE Perform canonical transformation based on Eq.\ref{eq10}: $y = P^\top \delta$,
where $P = [v_1, v_2, \dots, v_k]$ is the matrix of eigenvectors. Each component $y_i$ in $y$ is given by:
\[
y_i = v_i^\top \delta.
\]

\STATE \textbf{Step 4: Select $m$ short-axis eigenvalues and solve for $\delta$}
\STATE Select $y_i$ corresponding to the $m$ smallest eigenvalues $\lambda_i$ and formulate the optimization problem based on Eq.\ref{eq13}.

\STATE Use gradient descent to iteratively solve for $\delta$:
\[
\delta^{(t+1)} = \delta^{(t)} - \eta \nabla_{\delta} \|\delta\|^2,
\]
Where $\eta$ is the learning rate.

\STATE \textbf{Step 5: Compute quantization bit-width from $\delta$}
\STATE Using the optimized perturbation $\delta$, compute the quantization bit-width $b_i$ for each layer:
\[
b_i = \log_2\left(\frac{1}{|\delta_i| + \alpha}\right),
\]
where $\alpha \to 0$ is a small positive constant. \COMMENT {or quantization error mapping}

\STATE \textbf{Step 6: Output the quantization bit-width for each layer}
\RETURN $\{b_1, b_2, \dots, b_k\}$, where $k$ is the number of layers in the model.
\end{algorithmic}
\end{algorithm}

\vspace{-0.3cm}
\subsection{CET Algorithm for Quantization}
We attempt to apply the CET (Compression Error Theoretical) framework to model quantization. CET regards quantization as a process of introducing noise perturbation into the model parameters, where the error grows as the bit-width decreases. The detailed procedure is illustrated in Algorithm 1. CET utilizes the Lanczos algorithm to compute the eigenvalues and eigenvectors of the Hessian matrix. Subsequently, it formulates the indeterminate equation system based on Eq.\ref{eq13} and solves for the perturbation vector $\delta$ using gradient descent. Once $\delta$ is obtained, two approaches are proposed to calculate the bit-width: \textcircled{1} The first method directly computes the bit-width for each layer using the quantization formula provided in the algorithm. \textcircled{2} The second method compares the actual quantization loss under different bit-widths and selects the $\delta$ that aligns with the true quantization error.
CET combines both approaches to determine the optimal bit-width. When the first method fails to yield accurate results due to the impact of outliers, the second method is adopted to ensure reliability. Regarding time complexity, the primary computational bottleneck lies in the Lanczos algorithm, with a complexity of $\mathcal{O}(n^2)$. The remaining steps of CET are computationally efficient, with a complexity of $\mathcal{O}(n)$.

CET is highly extensible and can be applied to other model compression techniques. Most compression methods, such as quantization, decomposition, and parameter sharing, can be viewed as processes that introduce noise perturbations into the model parameters. CET provides a theoretical framework for analyzing these compression-induced errors and determines the optimal compression level in practice. As a general-purpose method, CET is not tailored to any specific compression technique but can be generalized across a wide range of compression methods.

\begin{figure}[!ht]
\centering
\vspace{-0.1cm}
\includegraphics[width=8.5cm,height=3.0cm]{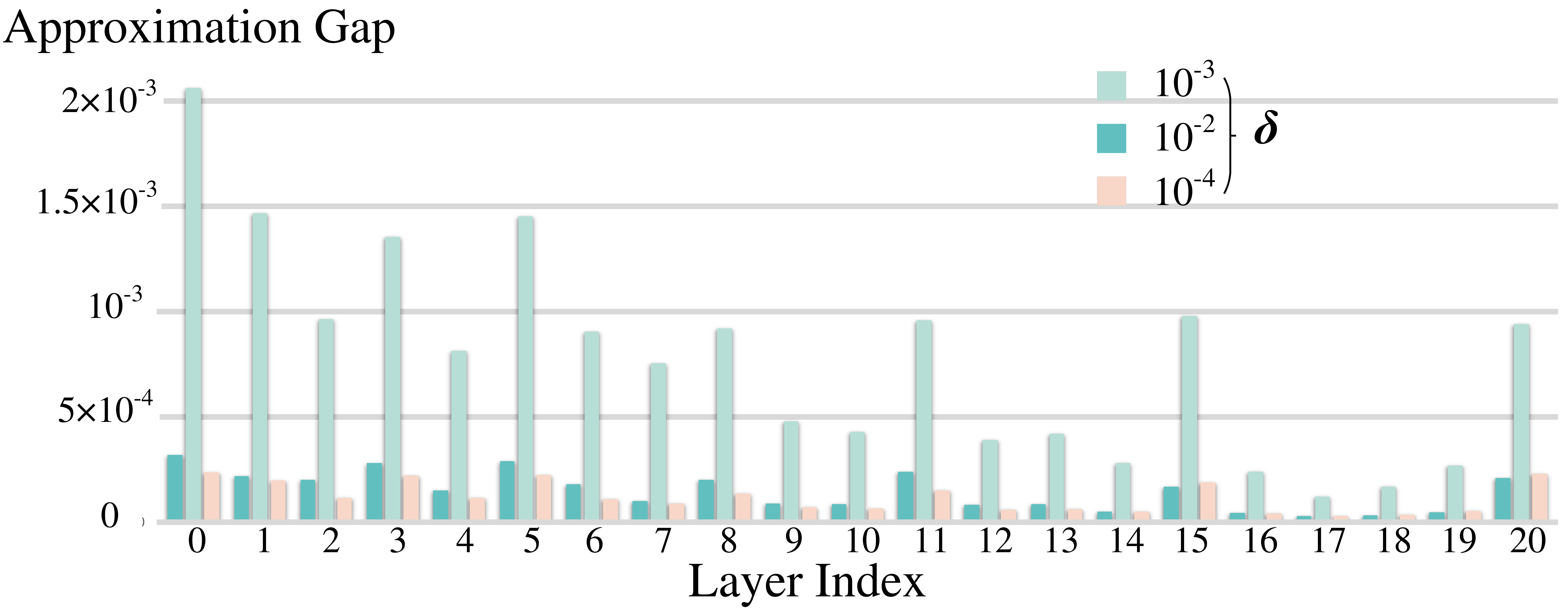}
\vspace{-0.9cm}
\caption{The gap between theory and practice after adding different perturbations to each layer. This ensures that the theoretical approximation can effectively represent the actual value in a sufficiently small neighborhood.}
\vspace{-0.4cm}
\label{gap}
\end{figure}
\section{Experiments}
\subsection{Datasets and Details}
We evaluate the performance of CET on various models on ImageNet. The ImageNet-1K dataset\cite{krizhevsky2017imagenet} consists of 1.28 million training and 50K validation images. ImageNet-1K is usually used as the benchmark for model compression. The calibration set is taken from the ImageNet validation set. SWAG dataset~\cite{zellers2018swag} consists of 113K multiple-choice questions about grounded situations. The Stanford Question Answering Dataset (SQuAD)~\cite{rajpurkar2016squad} is a collection of question-answer pairs derived from Wikipedia articles. In SQuAD, the correct answers to questions can be any sequence of tokens in the given text. MNLI~\cite{williams2017broad} is a dataset for natural language reasoning tasks. Its corpus is a collection of textual implication annotations of sentences through crowdsourcing. The task is to predict whether the premise sentence and the hypothesis sentence are logically compatible (entailment, contradiction, neutral). 

Following existing compression work, CET utilizes the Lanczos algorithm to compute the eigenvalues and eigenvectors of the Hessian matrix, with the maximum number of iterations set to 100. The Lanczos algorithm avoids explicitly constructing the Hessian matrix, efficiently approximating its eigenvalues and significantly reducing computational complexity. In the higher-order expansion (Eq.\ref{eq7}), when the actual perturbation error and the theoretically derived perturbation error are both less than $10^{-3}$, the discrepancy between theory and practice becomes negligible. For Eq.\ref{eq13}, it is formulated as an optimization problem where quantization error is minimized using the Adam optimizer. To ensure experimental consistency, all models use the same settings without tuning hyperparameters. The quantization bit-widths for each layer are calculated using the quantization algorithm \cite{zhang2024fp}, without fine-tuning or retraining. All experiments are conducted on two NVIDIA A800 GPUs, and the code is implemented in PyTorch, which will be made publicly available.

\begin{table*}[ht]
\centering
\vspace{-0.4cm}
\caption{Comparison with existing uniform quantization methods and the latest mixed precision methods. MP refers to mixed precision quantization, and we report the lowest bits used for weights and activations. w-ratio and a-ratio represent the weight and activation compression ratios, respectively.}
\renewcommand{\arraystretch}{1.2}
\setlength{\tabcolsep}{13.0pt}
\scalebox{0.83}{
\begin{tabular}{c|cccccccc} 
\hline
\textbf{Model}                        & \textbf{Method}                  & \textbf{Top-1/Full$\uparrow$}          & \textbf{w-bit}                  & \textbf{a-bit}                  & \textbf{w-ratio}             & \textbf{a-ratio}              & \textbf{Top-1/Quant$\uparrow$}         & \textbf{Top-1/Drop}           \\ 
\hline
\multirow{12}{*}{\textbf{ResNet-18 }} & LQ-Nets                          & 70.30                         & 3                               & 32                              & $\times$7.45                         & $\times$1.00                             & 69.30                         & -1.00                            \\
                                      & GPTQ                          & 69.76                         & 3                               & 32                              & $\times$10.66                          & $\times$1.00                             & 67.88                           & -1.88                          \\
                                      & MCKP                             & 69.76                        & 2$_{\textbf{MP}}$                             & 32                              & $\times$10.66                        & $\times$1.00                             & 69.50                         & -0.26                         \\
                                      & {\cellcolor[rgb]{0.941,1,1}}Ours &69.76 {\cellcolor[rgb]{0.941,1,1}} & {\cellcolor[rgb]{0.941,1,1}}2$_{\textbf{MP}}$ & {\cellcolor[rgb]{0.941,1,1}}32  &$\times$\textbf{11.02} {\cellcolor[rgb]{0.941,1,1}} & {\cellcolor[rgb]{0.941,1,1}}$\times$1.00 & 69.75{\cellcolor[rgb]{0.941,1,1}} & {\cellcolor[rgb]{0.941,1,1}} \textbf{-0.01}  \\
                                      & MCKP                             & 69.76                        & 2$_{\textbf{MP}}$                             & 8                               & $\times$10.66                        & $\times$4.00                             & 69.39                        & -0.37                         \\
                                      & {\cellcolor[rgb]{0.941,1,1}}Ours &69.76 {\cellcolor[rgb]{0.941,1,1}} & {\cellcolor[rgb]{0.941,1,1}}2$_{\textbf{MP}}$ & {\cellcolor[rgb]{0.941,1,1}}8   & $\times$\textbf{11.02} {\cellcolor[rgb]{0.941,1,1}} & {\cellcolor[rgb]{0.941,1,1}}$\times$4.00 &69.71 {\cellcolor[rgb]{0.941,1,1}} & {\cellcolor[rgb]{0.941,1,1}} \textbf{-0.05}  \\ 
\cline{2-9}
                                      & LQ-Nets                          & 70.30                         & 4                               & 4                               & $\times$6.10                          & $\times$7.98                          & 69.30                         & -1.00                            \\
                                      & DoReFa                           & 70.40                         & 5                               & 5                               & $\times$5.16                         & $\times$6.39                          & 68.40                         & -2.00                           \\
                                      & PACT                             & 70.40                         & 4                               & 4                               & $\times$6.10                          & $\times$7.98                          & 69.20                        & -1.20                          \\
                                      & MCKP                             & 69.76                        & 3$_{\textbf{MP}}$                             & 4$_{\textbf{MP}}$                             & $\times$8.32                         & $\times$8.00                             & 69.66                        & \textbf{-0.10 }                         \\
                                      & PTMQ                             & 71.00                        & 4$_{\textbf{MP}}$                            & 4$_{\textbf{MP}}$                                & $\times$8.00                        & $\times$8.00                             & 67.57                         & -3.43                      \\
                                      & HMQAT                &69.76              & 3$_{\textbf{MP}}$                       & 4                                                          & $\times$9.92                        & $\times$8.00                             & 68.73                         & -1.03                      \\                                      
                                      & {\cellcolor[rgb]{0.941,1,1}}Ours &69.76 {\cellcolor[rgb]{0.941,1,1}} & {\cellcolor[rgb]{0.941,1,1}}2$_{\textbf{MP}}$ & {\cellcolor[rgb]{0.941,1,1}}4$_{\textbf{MP}}$ & $\times $\textbf{11.02} {\cellcolor[rgb]{0.941,1,1}} & {\cellcolor[rgb]{0.941,1,1}}$\times$8.00 & 69.01{\cellcolor[rgb]{0.941,1,1}} & {\cellcolor[rgb]{0.941,1,1}}-0.65 \\ 
\hline
\multirow{2}{*}{\textbf{ResNet-34}}& {}Ours &73.22  & 2$_{\textbf{MP}}$ & 4$_{\textbf{MP}}$ & $\times $\textbf{13.44}  &$\times$8.00 & 72.60 & -0.62            \\
& {\cellcolor[rgb]{0.941,1,1}}Ours & 73.22 {\cellcolor[rgb]{0.941,1,1}} & {\cellcolor[rgb]{0.941,1,1}}2$_{\textbf{MP}}$ & {\cellcolor[rgb]{0.941,1,1}}32 & $\times $\textbf{10.96} {\cellcolor[rgb]{0.941,1,1}} & {\cellcolor[rgb]{0.941,1,1}}$\times$1.00 & 73.33{\cellcolor[rgb]{0.941,1,1}} & {\cellcolor[rgb]{0.941,1,1}} \textbf{+0.11} \\

\hline
\multirow{12}{*}{\textbf{ResNet-50 }}
                                    
                                      & DoReFa                           & 76.90                         & 4                               & 4                               & $\times$5.11                         & $\times$7.99                          & 71.40                         & -5.50                          \\
                                      & PACT                             & 76.90                         & 32                              & 4                               & $\times$1.00                            & $\times$7.99                          & 75.90                         & -1.00                            \\
                                      & AutoQ                            & 74.80                         & MP                              & MP                             & $\times$10.26                        & $\times$7.96                          & 72.51                        & -2.29                         \\
                                      & HAWQ                             & 77.39                        & 2$_{\textbf{MP}}$                             & 4$_{\textbf{MP}}$                             & $\times$12.28                        & $\times$8.00                             & 75.48                        & -1.91                         \\
                                      & HAWQ-v2                          & 77.39                        & 2$_{\textbf{MP}}$                             & 4$_{\textbf{MP}}$                             & $\times$12.24                        & $\times$8.00                             & 75.76                        & -1.63                         \\
                                      
                                      & MCKP                             & 76.13                        & 2$_{\textbf{MP}}$                             & 4$_{\textbf{MP}}$                             & $\times$12.24                        & $\times$8.00                             & 75.28                        & \textbf{-0.85}                         \\
                                    & PTMQ                             & 76.80                        & 4$_{\textbf{MP}}$                            & 4$_{\textbf{MP}}$                                & $\times$8.00                        & $\times$8.00                             & 73.93                         & -2.87                      \\
                                      & {\cellcolor[rgb]{0.941,1,1}}Ours & 76.12 {\cellcolor[rgb]{0.941,1,1}} & {\cellcolor[rgb]{0.941,1,1}}2$_{\textbf{MP}}$ & {\cellcolor[rgb]{0.941,1,1}}4$_{\textbf{MP}}$ & $\times$\textbf{13.53} {\cellcolor[rgb]{0.941,1,1}} & {\cellcolor[rgb]{0.941,1,1}} $\times$8.00  & {\cellcolor[rgb]{0.941,1,1}} 75.13 & {\cellcolor[rgb]{0.941,1,1}} -0.99  \\
                                      & HAQ                              & 76.15                        & MP                             & 32                              & $\times$10.57                        & $\times$1.00                             & 75.30                         & -0.85                         \\
                                      & OBQ                              & 76.13                        & 3                             & 32                              & $\times$10.66                        & $\times$1.00                             & 75.24                         & -0.89                        \\
                                    & GPTQ                             & 76.13                        & 3                             & 32                              & $\times$10.66                        & $\times$1.00                             & 74.87                         & -1.26                       \\
                                        & {\cellcolor[rgb]{0.941,1,1}}Ours & 76.12 {\cellcolor[rgb]{0.941,1,1}} & {\cellcolor[rgb]{0.941,1,1}}2$_{\textbf{MP}}$ & {\cellcolor[rgb]{0.941,1,1}}32 & $\times$\textbf{12.74} {\cellcolor[rgb]{0.941,1,1}} & {\cellcolor[rgb]{0.941,1,1}} $\times$1.00  & {\cellcolor[rgb]{0.941,1,1}} 76.09 & {\cellcolor[rgb]{0.941,1,1}} \textbf{-0.03 } \\

\hline
\end{tabular}} \label{tab1}
\vspace{-0.3cm}
\end{table*}

\begin{table}
\centering
\vspace{-0.3cm}
\caption{Comparison with existing mixed precision methods on the MobileNet-V2 model.}
\renewcommand{\arraystretch}{1.06}
\setlength{\tabcolsep}{2.8pt}
\scalebox{0.83}{
\begin{tabular}{ccccccc} 
\hline
\textbf{Method}                & \textbf{w-bit} & \textbf{a-bit} & \textbf{w-ratio} & \textbf{a-ratio} & \textbf{Top-1/Quant} & \textbf{Top-1/Drop}  \\ 
\hline
DC                             & MP             & 32             & $\times$13.93            & $\times$1                & 58.07                & -13.8                \\
HAQ                            & MP             & 32             & $\times$14.07            & $\times$1                & 66.75                & -5.12                \\
MCKP                           & 2MP            & 8              &$\times$13.99            & $\times$4                & 68.52                & -3.36                \\
\rowcolor[rgb]{0.949,1,1} Ours & 2MP            & 32             &$\times$14.99                 &  $\times$1               &  70.14                    &  \textbf{ -1.69 }                  \\
\rowcolor[rgb]{0.949,1,1} Ours & 2MP            & 8              &  $\times$14.99                &   $\times$4              &    69.66                   &  \textbf{-2.17}                    \\
\hline
\end{tabular}} \label{tab2}
\vspace{-0.7cm}
\end{table}

\begin{table*}[!ht]
\centering
\vspace{-0.3cm}
\caption{Performance of the BERT\_base model on multiple language processing datasets when weights are quantized to $4_{MP}$ bits.}
\renewcommand{\arraystretch}{1.06}
\setlength{\tabcolsep}{17.5pt}
\scalebox{0.83}{
\begin{tabular}{c|ccccccc}
\hline
\multicolumn{1}{c|}{\multirow{2}{*}{Method}} & \multicolumn{3}{c}{SQuAD1.1} & \multicolumn{2}{c}{MNLI} & SWAG  & w-ratio \\ \cline{2-8} 
\multicolumn{1}{c|}{}                        & Acc on Val$\uparrow$  & F1$\uparrow$     & EM$\uparrow$    & Acc on Val$\uparrow$ & Acc on Test$\uparrow$ & Acc$\uparrow$   &         \\ \hline
Full Prec.                                   & 85.74       & 88.42  & 80.89 & 82.77      & 84.57       & 79.11 &  $\times$1.00       \\
ACIQ                                         & 80.11       & 84.18  & 77.34 & 76.37      & 79.64       & 76.52 &$\times$6.07   \\
Zhang et.al.                                 & 85.67       & \textbf{88.16}  & 80.42 & \textbf{82.78}      & \textbf{83.92}       & 78.30  & $\times$1.58   \\
\rowcolor[rgb]{0.949,1,1} Ours                                         & \textbf{85.69}      & \textbf{88.16}  & \textbf{80.57} & 82.75      & 82.96       & \textbf{79.04} & \textbf{$\times$7.71}   \\ \hline
\end{tabular}} \label{tab3}
\vspace{-0.5cm}
\end{table*}
\vspace{-0.1cm}
\subsection{Comparison}
\textbf{Algorithm Accuracy.} As shown in Table \ref{tab1}, we conducted a comprehensive comparison between CET and existing quantization methods \cite{lin2017towards, zhang2018lq, zhou2016dorefa, choi2018pact, han2015deep, huang2025hessian,xu2024ptmq, frantar2022optimal}. For ResNet-18 \cite{he2016deep}, CET achieved nearly lossless weight quantization, with model performance degrading by only 0.01\%, while significantly reducing the model size, achieving a compression ratio of over 11$\times$. This demonstrates the effectiveness of CET's geometric analysis based on second-order information, enabling precise weight compression along the long-axis direction and selecting appropriate error tolerance for each layer. Furthermore, to investigate the impact of activation compression on weight quantization, we performed quantization on activations with different bit widths (8-bit and 4-bit). The results show that CET can maximize model compression while maintaining accuracy. Notably, CET does not rely on fine-tuning but directly determines the optimal bit-width allocation, which distinguishes it from methods like MCKP that require fine-tuning.

For ResNet-50, CET demonstrates significant performance advantages under higher compression rates. Compared with HAWQ and HAWQ-V2, which also utilize second-order information, CET achieves smaller accuracy degradation under more extreme compression scenarios (-0.99\% vs. -1.91\%). Additionally, compared with HAQ, which searches for optimal bit-width allocation via reinforcement learning, CET achieves a higher weight compression rate with minimal accuracy degradation and significantly lower computational cost. For example, after achieving a 12.74$\times$ weight compression, CET reduces model accuracy by only 0.034\%, effectively realizing lossless compression.
Compared with MCKP, CET achieves minimal performance degradation even at a higher compression rate (13.53$\times$), while MCKP experiences similar performance degradation at a lower compression rate (12.24$\times$). This highlights CET's stronger robustness and generalization capabilities under higher compression demands, achieving superior performance retention at extreme compression ratios.

For ResNet-34, CET even improves model accuracy while achieving nearly 11$\times$ compression. This is due to ResNet-34's second-order geometric property, where the Hessian matrix at the convergence point is indefinite. CET performs quantization along the eigenvector direction corresponding to negative eigenvalues, reducing the loss and resulting in higher accuracy for the quantized model compared to the original. CET rigorously analyzes the Hessian matrix's shape using algebraic geometry, enabling it to select the optimal quantization direction.
In Table \ref{tab2}, finally, we further evaluate CET on the lightweight and efficient MobileNet-V2 architecture. The results show that CET achieves significant performance improvements even at higher compression rates. This further validates the rationality and broad applicability of CET's second-order geometric analysis.

Table \ref{tab3} presents CET's performance across various NLP datasets. With 4-bit mixed-precision weight quantization, CET achieves accuracy comparable to the original model. While Zhang et al.’s method shows a slight advantage on the MNLI dataset, its low compression rate makes it difficult to maintain stable accuracy under high compression.

\textbf{Algorithm Efficiency.} The CET algorithm demonstrates high computational efficiency when solving indeterminate equations using gradient descent. Experiments show that it converges in approximately 2000 iterations, taking only a few minutes. Moreover, the main computational complexity and time consumption of the CET algorithm stem from the Lanzcos algorithm. This is because it requires calculating the Hessian information for each sample in the calibration set. This time consumption characteristic aligns with the computational speed of existing algorithms.

\vspace{-0.2cm}
\subsection{Ablation}
\textbf{Differential Expansion. }
Differential expansion ablation is to explore the effective neighborhood of CET at each layer. Differential expansion requires ensuring that the neighborhood around the expansion point is sufficiently small to guarantee that the theoretical approximation effectively represents the actual values within this neighborhood. Figure \ref{gap} shows the discrepancies between theoretical and actual values under different perturbation errors introduced into each layer. Taking ResNet-18 as an example, the gap between theoretical and actual values is minimal across layers, which validates the effectiveness of CET. Additionally, error perturbation experiments reveal that different layers exhibit distinct levels of error tolerance.

\begin{figure}[!ht]
\centering
\vspace{-0.2cm}
\includegraphics[width=5.9cm,height=4.0cm]{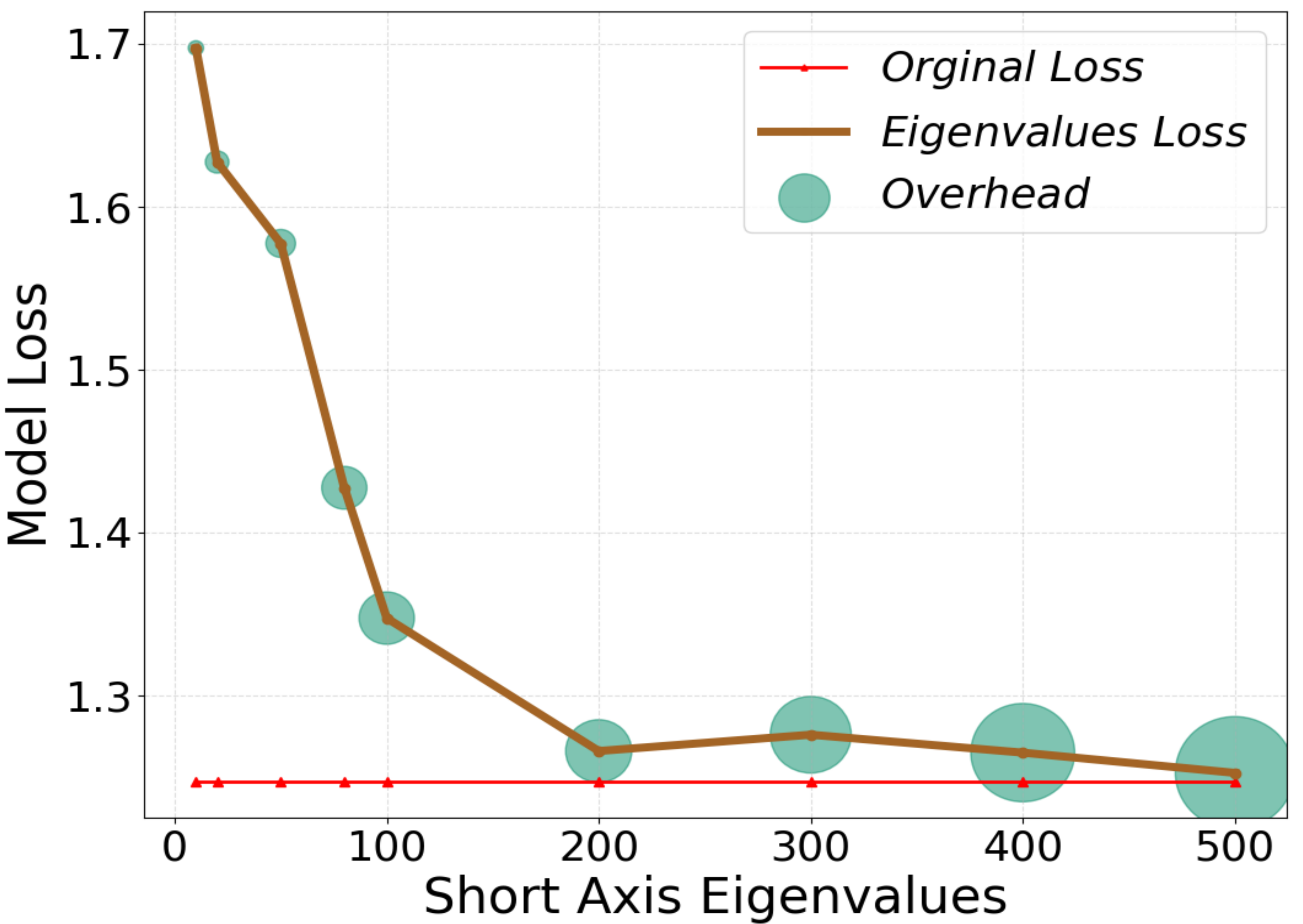}
\vspace{-0.3cm}
\caption{As the number of short-axis eigenvalues increases, the number of equations in the underdetermined system of Eq.\ref{eq13} also grows, leading to a decrease in model loss. However, this increase is accompanied by a corresponding rise in computational overhead (the area of the circle increases).}
\vspace{-0.3cm}
\label{evalue}
\end{figure}

\textbf{Gradient and Short Axis Eigenvalue. }
Theoretically, the gradient of a well-trained model approach zero. Practical inspection reveals the first-order gradient values are small, with most being less than $10^{-5}$. As a result, the influence of the first-order term on the loss can be safely neglected. Given the high dimensionality of the Hessian matrix, computing all eigenvalues directly is infeasible. To balance memory consumption and computational efficiency, CET truncates the short-axis eigenvalues, calculating only a subset of them. Figure \ref{evalue} illustrates the impact of truncated short-axis eigenvalues on the loss. When the number of eigenvalues is set to 50, 100, 200, and 500, the loss shows a decreasing trend. However, beyond 500 eigenvalues, the rate of loss reduction diminishes sharply, while the computational cost rises sharply. Consequently, we select 200 short-axis eigenvalues as a compromise in implementation.

\vspace{-0.2cm}
\subsection{Discussion}
\textbf{Limitations.} Ideally, the CET requires precise Hessian information to ensure accurate analysis of quantization errors. However, since directly computing the full Hessian matrix is infeasible in practice, we approximate it using the Lanzcos algorithm. While this approximation introduces some degree of error, our perturbation analysis of the Lanzcos algorithm reveals that it has a strong error tolerance. The approximate Hessian information obtained is robust in most cases, ensuring the practical applicability of the CET.

\textbf{Generality.} The CET framework is algorithm-agnostic and features strong interpretability and adaptability. It does not rely on any specific compression strategy but rather provides a universal theoretical foundation.
Consequently, CET can be seamlessly applied to various compression strategies. Offering an effective geometric analysis framework, it helps these methods identify optimal compression configurations.
\vspace{-0.5cm}
\section{Conclusion}
We proposed a Compression Error Theory (CET) framework designed to determine the optimal compression level for each layer of a model. During quantization, CET reconstructs the quadratic form of quantization error into different geometric structures and reformulates the optimization process as a complementary problem to solve the error subspace. Theoretical analysis shows that constructing the quantization subspace along the long axis minimizes the impact on model performance. Extensive experimental results further validate the effectiveness of CET. In the future, CET is a general compression theory framework that can be extended to other compression methods, such as weight decomposition.

\nocite{langley00}

\bibliography{example_paper}
\bibliographystyle{icml2025}




\end{document}